%% file: main.tex
\documentclass{article}

\usepackage[preprint]{neurips_2026}

\usepackage[utf8]{inputenc}
\usepackage[T1]{fontenc}
\usepackage{amsmath,amssymb,amsfonts}
\usepackage{booktabs}
\usepackage{makecell}
\usepackage{multirow}
\usepackage{tabularx}
\usepackage{graphicx}
\usepackage{placeins}
\usepackage{microtype}
\usepackage{xcolor}
\usepackage{url}

\definecolor{mycolor_1}{RGB}{140, 27, 19}
\definecolor{mycolor_1}{RGB}{5, 170, 75}
\definecolor{darkblue}{RGB}{0,45,110}

\usepackage{hyperref}
\hypersetup{
    colorlinks=true,
    linkcolor=blue,
    filecolor=magenta,      
    urlcolor=blue,
    citecolor=darkblue
}       

\title{Where Does Long-Context Supervision Actually Go? Effective-Context Exposure Balancing}

\author{%
  \bfseries Jinchang Zhu$^{1,a}$, Jindong Li$^{1}$, Chengyu Zou, Rong Fu, \\
  \bfseries Chao Wang$^{2}$, Haowei He$^{2}$, Menglin Yang$^{1,\dagger,b}$ \\[0.3em]
  \mdseries $^{1}$The Hong Kong University of Science and Technology (Guangzhou) \\
  \mdseries $^{2}$Institute of Artificial Intelligence (TeleAI), China Telecom \\[0.2em]
  \mdseries $^{a}$\texttt{jzhu997@connect.hkust-gz.edu.cn} \quad $^{b}$\texttt{menglinyang@hkust-gz.edu.cn} \\[0.2em]
  \mdseries $^{\dagger}$Corresponding author.
}

\newcommand{\method}{\textbf{EXACT}}
\newcommand{\ctx}{\ell}

\begin{document}

\maketitle

\input{sections/00_abstract}
\input{sections/01_introduction}

\input{sections/02_related_work}
\input{sections/03_method}
\input{sections/04_experimental_setup}
\input{sections/experiments/01_qwen05_cpt}
\input{sections/experiments/03_preservation_and_mechanism}
\input{sections/05_analysis}
\input{sections/06_limitations}
\input{sections/07_conclusion}

\clearpage

\bibliographystyle{plainnat}
\bibliography{references}

\clearpage

\appendix
\input{sections/09_appendix}

\end{document}

%% file: sections/00_abstract.tex
\begin{abstract}
Long-context adaptation is commonly treated as window scaling: increase the training sequence length, continue pretraining, and expect long-context behavior to improve. This view \textbf{\textit{misses a token-level supervision mismatch}}: longer windows provide longer possible contexts, while causal language modeling delivers supervision one target token at a time. In packed training with document-boundary masking, each target token has its own \textbf{\textit{effective left context}}, the visible same-document prefix at prediction time, and \textbf{\textit{the resulting exposure histogram remains heavily concentrated in short and medium regimes}}. We introduce \method{} (Effective-context Allocation for Context Training), a supervision-allocation objective that measures this exposure distribution, assigns a fixed extra supervision weight to long effective-context targets, and distributes that extra weight by inverse frequency within the long-context tail. Across seven Qwen and LLaMA continued-pretraining (CPT) configurations, \method{} improves all twenty-eight trained/extrapolated NoLiMa and RULER comparisons in the main table. On Qwen2.5-0.5B, it improves NoLiMa by +10.09 points in trained lengths and +5.34 in extrapolated lengths, and improves RULER by +10.69 and +5.55. On LLaMA-3.2-3B, it improves RULER by +17.91 and +16.11 in trained and extrapolated regimes. Standard QA and reasoning are preserved, with a +0.24 macro change across MMLU, ARC-C, HellaSwag, WinoGrande, PIQA, and GSM8K. A distance-resolved evidence-ablation probe further shows that the gains arise in the expected regime: when relevant evidence is placed thousands of tokens away from the query, \method{} makes the model more likely to prefer the correct answer, while short-distance cases remain close to standard CPT. These results support a supervision-centric thesis: \textbf{\textit{long-context adaptation depends not only on what context a model can process, but also on how strongly training supervises predictions made under long effective context.}}
\end{abstract}

%% file: sections/01_introduction.tex
\section{Introduction}

Long-context language models are usually adapted by increasing the training window \citep{chen2023longlora,bai2024longalign,fu2024dataengineering,ding2024longrope}. This practice treats sequence length as the unit of long-context supervision. Sequence length is a capacity variable: it sets the maximum span a model may process in a forward pass. Supervision length is a token-level variable: it is determined by how much usable left context each supervised target actually has. A 4K or 8K packed sequence can contain many targets whose same-document left context is short, especially under document-boundary masking and realistic document length mixtures.

The right unit of analysis is therefore the target token. For a token at position $i$, its effective left context $\ctx_i$ is the number of previous same-document tokens visible when predicting that token. This quantity depends on causal masking, document boundaries, packing layout, and the length distribution of the selected documents. Data selection decides which documents enter the stream and how much long text is available. The training objective decides how supervision is allocated across the tokens produced by that stream. These two layers interact, but they are distinct. Even with a long-window corpus, the training signal can remain concentrated on short and medium effective-context targets.

\begin{figure}[t]
\centering
\includegraphics[width=0.88\linewidth]{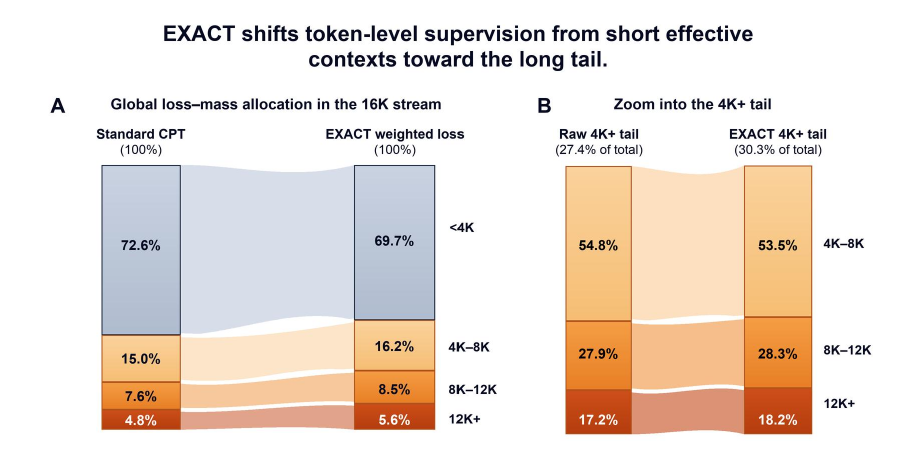}
\caption{Standard 16K CPT still places most token-level supervision below 4K effective context. \method{} keeps the packed stream fixed but changes per-token loss weights; when weighted objective contributions are normalized for visualization, the relative loss mass shifts into the 4K+ tail, and within that tail farther effective-context buckets receive a larger share.}
\label{fig:where-supervision-goes}
\end{figure}

Figure~\ref{fig:where-supervision-goes} makes the mismatch visible. The packed sequence may span 16K tokens, yet each supervised target sees only the prefix of its own document segment. In the measured 16K stream, 72.6\% of standard-CPT loss mass still lands below 4K effective context, leaving only 27.4\% in the 4K+ tail. \method{} keeps the stream fixed and changes the supervision allocation: the 4K+ tail rises to 30.3\%, and the farther buckets receive a larger share within that tail. This is the supervision-allocation bottleneck targeted by this paper: \textit{\textbf{long-context adaptation pays for long windows, while a large share of training signal still lands on targets with short or medium effective context}}.

We introduce \method{}, Effective-context Allocation for Context Training, to make this hidden allocation explicit. \method{} computes each supervised token's \textbf{\textit{effective left-context length}}, maps it into logarithmic buckets, and assigns fixed extra supervision weight to the long-context tail. Within that tail, the added weight is distributed by inverse bucket frequency, so rare long-effective-context targets become first-class training events rather than being washed out by the much larger population of short-context targets.

The empirical pattern follows this mechanism. Across seven Qwen and LLaMA continued-pretraining configurations, \method{} improves every trained and extrapolated NoLiMa/RULER aggregate in the main table. On Qwen2.5-0.5B 4K CPT, NoLiMa rises by +10.09 points in trained lengths and +5.34 in extrapolated lengths, while RULER rises by +10.69 and +5.55. On LLaMA-3.2-3B staged 4K$\rightarrow$8K CPT, RULER improves by +17.91 and +16.11 in trained and extrapolated regimes. Standard QA is preserved, with a +0.24 macro change across MMLU, ARC-C, HellaSwag, WinoGrande, PIQA, and GSM8K. The strongest normalized gains occur in regimes where long effective-context supervision should matter most. A distance-resolved evidence probe then shows the same shape inside the model: \method{} leaves short distances near standard long-window CPT parity, but opens positive context-induced margin gains from 4K to 16K. The preservation and probe results show why the long-context scores move without sacrificing ordinary QA competence: distant evidence becomes more usable.

Our contributions are summarized as follows:
\begin{itemize}
    \item We identify \textbf{effective-context exposure imbalance} as a measurable supervision-allocation bottleneck in packed causal language modeling.
    \item We propose \method, a \textbf{token-level objective} that assigns fixed extra supervision weight to the long-context tail and distributes it by \textbf{inverse effective-context bucket frequency}.
    \item We show consistent gains on official generation-based NoLiMa and RULER evaluations across seven Qwen and LLaMA continued-pretraining configurations, with standard QA preserved and a distance-resolved evidence probe confirming stronger conversion of far context into answer likelihood.
\end{itemize}

%% file: sections/02_related_work.tex
\section{Related Work}

\paragraph{Long-context capacity and positional range.}
A large body of work extends the amount of context that language models can process. Transformer recurrence and memory mechanisms extend usable history across segments \citep{dai2019transformerxl,rae2019compressive}, while sparse or approximate attention reduces long-sequence cost through structured attention patterns or kernel approximations \citep{child2019sparse,beltagy2020longformer,zaheer2020bigbird,kitaev2020reformer,choromanski2020rethinking}. Efficient exact-attention and KV-cache methods improve the practicality of long-window training and inference \citep{dao2022flashattention,dao2023flashattention,liu2023ringattention,ainslie2023gqa}, and architectures such as LongNet and Infini-attention explore extremely long or unbounded contexts \citep{ding2023longnet,munkhdalai2024infiniattention}. A complementary line extends the positional range of pretrained models through RoPE, ALiBi, interpolation, or rescaling \citep{su2024roformer,press2021train,chen2023extending,peng2023yarn,ding2024longrope,jin2024selfextend,xiao2023efficient,han2024lminfinite}. These methods expand what context can be processed and where models can operate. \method{} studies a different training-objective question: once long windows are available, how much supervision reaches targets predicted under long same-document effective context?

\paragraph{Long-context continued pretraining and data supply.}
Continued pretraining and long-context fine-tuning are standard tools for adapting LLMs to longer inputs. LongLoRA makes long-context fine-tuning efficient through sparse attention and parameter-efficient adaptation \citep{chen2023longlora}. LongAlign provides a long-context alignment recipe involving instruction data, packing, sorted batching, and loss weighting \citep{bai2024longalign}. Data-engineering work for 128K context shows that long-context CPT depends on token budget, domain balance, and length upsampling \citep{fu2024dataengineering}. Public Qwen and LLaMA models provide common bases for these adaptation studies \citep{touvron2023llama2,grattafiori2024llama,2025_arXiv_Qwen2.5_Qwen2.5-Technical-Report}. These works control the supply side: which documents enter training, how examples are packed, and how much token budget is spent at each window length. \method{} keeps the stream fixed and changes how supervision is allocated across the target tokens produced by that stream.

\paragraph{Long-context evaluation and context use.}
Evaluation work shows that available context is not the same as used context. Lost in the Middle demonstrates that models can fail to use relevant information depending on its position in the prompt \citep{liu2023lostmiddle}. LongBench, L-Eval, BAMBOO, LooGLE, InfiniteBench, HELMET, ZeroSCROLLS, and NoCha broaden long-context evaluation across QA, summarization, retrieval, code, synthetic tasks, and application-centric settings \citep{bai2024longbench,an2024leval,dong2024bamboo,li2024loogle,zhang2024infinitebench,yen2024helmet,shaham2023zeroscrolls,karpinska2024nocha}. RULER adds configurable retrieval, tracking, aggregation, and QA tasks beyond vanilla needle retrieval \citep{hsieh2024ruler,xue2024question}. NoLiMa removes lexical-overlap shortcuts and requires latent association between the query and the evidence \citep{modarressi2025nolima,xue2025structcoh}. We use generation-based NoLiMa and RULER because they test whether available long context is converted into answer production, matching our supervision-allocation hypothesis.

\paragraph{Token-level supervision and loss allocation.}
Non-uniform training signal is a long-standing idea, appearing in curriculum learning, self-paced learning, hard-example mining, focal loss, dataset cartography, and data-mixture optimization \citep{bengio2009curriculum,kumar2010selfpaced,shrivastava2016ohem,lin2017focalloss,swayamdipta2020datasetcartography,xie2023doremi}. Recent language-modeling work also challenges uniform token training: Rho-1 uses a reference model to select useful tokens for pretraining \citep{lin2024rho1}, long-context perplexity analyses question whether aggregate perplexity reflects long-context ability \citep{fang2024longppl}, and Token Weighting for Long-Range Language Modeling uses short-context versus long-context model confidence to assign token weights \citep{helm2025tokenweighting}. \method{} shares the view that token-level loss allocation matters, but targets a structural exposure variable rather than semantic token importance: the same-document effective left-context length of each supervised target in a packed causal-LM stream.

\FloatBarrier

%% file: sections/03_method.tex
\section{\method{}: Effective-Context Allocation}
\label{sec:method}

\begin{figure}[t]
\centering
\includegraphics[width=0.99\linewidth]{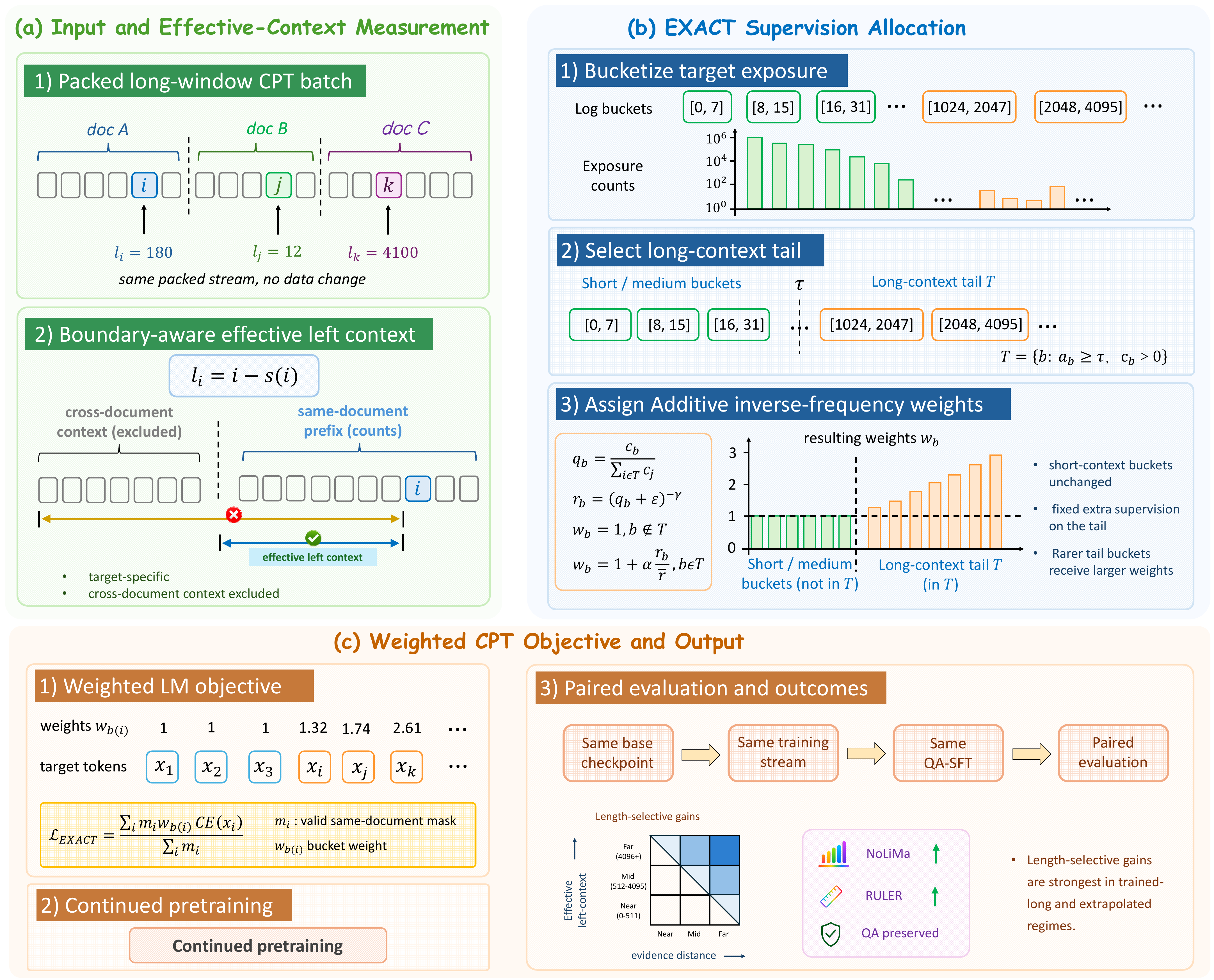}
\caption{The overview of \method{}.}
\label{fig:framework}
\end{figure}

\subsection{Effective left context}

Consider a packed causal language-modeling batch. For each target token position $i$, define its effective left-context length $\ctx_i$ as the number of previous tokens that are both causally visible and inside the same document segment:
\begin{equation}
    \ctx_i = i - s(i),
\end{equation}
where $s(i)$ is the start position of the current document segment inside the packed sequence. Cross-document tokens are excluded from $\ctx_i$ under document-boundary masking. This quantity is target-specific: a sequence of length $L$ contains targets with $\ctx_i$ ranging from zero to at most $L-1$, with most targets inheriting their context length from their position inside a document segment.

\subsection{Exposure buckets and token weights}

\method{} groups target tokens by logarithmic effective-context buckets:
\begin{equation}
    [0,7], [8,15], [16,31], \ldots, [1024,2047], [2048,4095], \ldots
\end{equation}
Let $b(i)$ be the bucket of target token $i$, and let $c_b$ denote the number of supervised tokens in bucket $b$, estimated from the training stream. These bucket statistics summarize how the chosen corpus, packing policy, and document-boundary mask translate into token-level supervision. They also expose the gap between nominal window length and the actual allocation of training signal.

The reported \method{} objective uses additive directional inverse-frequency allocation. Given a long-context threshold $\tau$, let $\mathcal{T}=\{b: a_b \geq \tau, c_b>0\}$ be the long-context tail, where $a_b$ is the lower bound of bucket $b$. Inside this tail, \method{} forms the tail-local bucket probability, inverse-frequency score, and frequency-weighted normalizer:
\begin{equation}
    q_b = \frac{c_b}{\sum_{j\in\mathcal{T}} c_j},
    \qquad
    r_b = (q_b + \epsilon)^{-\gamma},
    \qquad
    \bar{r} = \sum_{j\in\mathcal{T}} q_j r_j .
\end{equation}
The bucket weight is then
\begin{equation}
    w_b =
    \begin{cases}
        1, & b \notin \mathcal{T}, \\
        1 + \alpha \frac{r_b}{\bar{r}}, & b \in \mathcal{T}.
    \end{cases}
\end{equation}
This construction leaves short-context buckets unchanged and assigns the long-context tail an average extra weight of $\alpha$:
\begin{equation}
    \sum_{b\in\mathcal{T}} q_b (w_b - 1) = \alpha .
\end{equation}
The language-modeling objective becomes
\begin{equation}
    \mathcal{L}_{\method} =
    \frac{\sum_i m_i\, w_{b(i)}\, \mathrm{CE}(x_i)}
         {\sum_i m_i},
\end{equation}
where $m_i$ is the standard loss mask and $\mathrm{CE}(x_i)$ is the next-token cross-entropy at position $i$. The threshold $\tau$ is set relative to the training stage so that the extra loss weight is placed on the effective-context regime that the stage is meant to adapt. The hyperparameter $\alpha$ fixes the average extra weight assigned inside the long tail, while $\gamma$ controls how sharply that added weight moves toward rarer long-context buckets.

The objective targets a concrete behavioral variable: whether distant evidence changes the model's preference for the correct answer. We measure this with a paired NoLiMa evidence-sensitivity diagnostic. For each prompt, we compare the gold-answer margin under the original context with the margin under a counterfactual context where the supporting evidence is replaced by a same-type distractor. Their difference gives a context-induced answer margin $G$, which measures how strongly the available evidence is converted into answer likelihood at a given distance.

\begin{figure}[t]
\centering
\includegraphics[width=\linewidth]{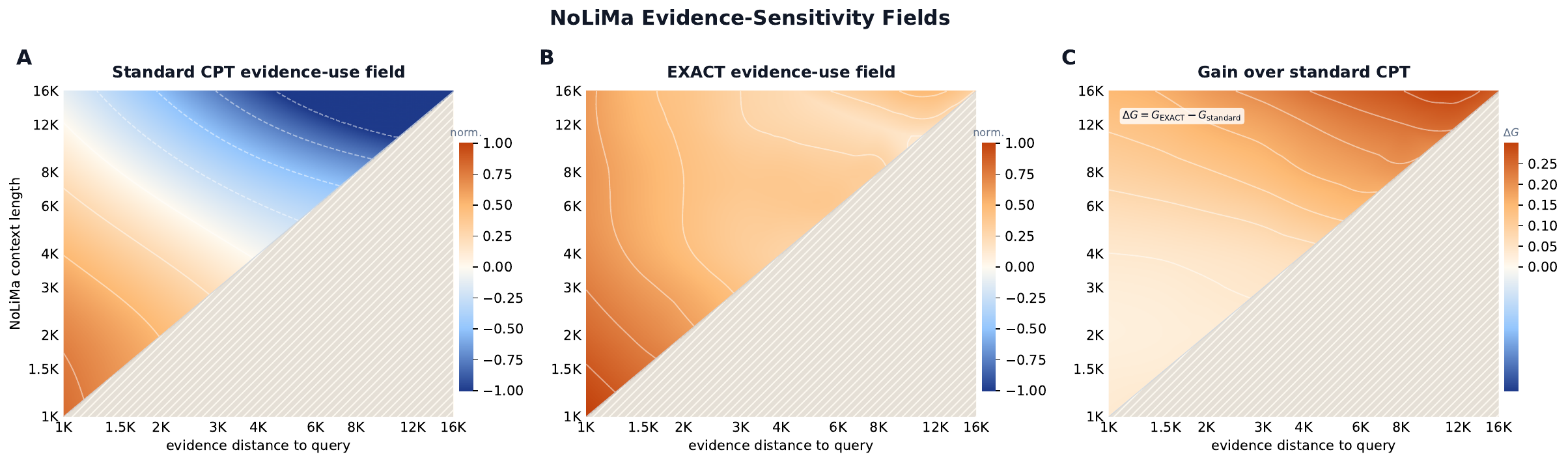}
\caption{NoLiMa evidence-sensitivity fields on paired LLaMA-3.1-8B 16K continuation checkpoints. Each cell groups prompts by context length and evidence distance to the query; the lower-right geometry where distance exceeds context length is not constructible. For each cell, we first compute the mean context-induced answer margin $G$: the gold-answer margin in the original prompt minus the margin under a counterfactual prompt whose supporting evidence is replaced by a same-type distractor. Panels A and B visualize the resulting $G$ fields under a shared robust display normalization for readability; Panel C reports the unnormalized gain $\Delta G = G_{\method} - G_{\mathrm{standard}}$. Gains concentrate in the far-evidence, long-context regime targeted by the objective.}
\label{fig:exposure-reallocation}
\end{figure}

Figure~\ref{fig:exposure-reallocation} shows the empirical footprint of the allocation change. Standard long-window CPT leaves the far/long corner weak: when the relevant evidence is distant and the context is long, the evidence contributes little to the answer margin. \method{} changes the field under the same prompt grid and visual scale. The strongest gains appear in the upper-right regime, precisely where standard CPT supplies relatively little long-effective-context supervision and where the reweighted tail should matter most.

\subsection{Expected behavior}

The same mechanism predicts length-selective improvement in downstream evaluation. Benchmarks that depend on trained-long and extrapolated long-context use should gain more than short-control regimes, because their success requires predictions conditioned on long effective left context. Ordinary short-context QA should be preserved because the objective still trains on all supervised targets while strengthening long-effective-context targets. Figure~\ref{fig:exposure-reallocation} makes this signature visible at the evidence-use level, and the following experiments test whether the benchmark pattern follows the same geometry.

%% file: sections/04_experimental_setup.tex
\section{Experimental Setup}
\label{sec:experimental-setup}

\paragraph{Models.}
We evaluate \method{} on Qwen2.5-0.5B, Qwen2.5-1.5B, Qwen2.5-3B, Qwen2.5-7B, LLaMA-3.2-1B, LLaMA-3.2-3B, and LLaMA-3.1-8B base models \citep{2025_arXiv_Qwen2.5_Qwen2.5-Technical-Report,touvron2023llama2,grattafiori2024llama}. Each model is adapted by continued pretraining (CPT) followed by the same question-answer supervised fine-tuning (QA-SFT) procedure on Databricks Dolly-15K instruction-response data \citep{DatabricksBlog2023DollyV2} for generation-based evaluation. The QA-SFT stage is held fixed across standard long-window CPT and \method{}.

\paragraph{Continued pretraining.}
Following common long-context extension practice, we use staged CPT before QA-SFT and keep each standard-CPT/\method{} comparison paired in initialization, token budget, data stream, QA-SFT corpus, and evaluation protocol. Appendix~\ref{app:cpt-corpus} reports the stage and corpus details. This pairing isolates effective-context supervision allocation. Unless otherwise stated, \method{} uses $\alpha=0.15$, $\gamma=0.5$, and $\epsilon=10^{-4}$; the long-tail threshold is set to $\tau=1$K, $2$K, and $4$K for the 4K, 8K, and 16K CPT stages, respectively.

\paragraph{Benchmarks.}
Long-context performance is measured with official generation-based NoLiMa and RULER evaluation. NoLiMa is reported over six public task sets: base, hard, distractor, multiple choice, only-direct, and chain-of-thought-style prompts. RULER is reported over 11 synthetic retrieval and tracking tasks with 500 examples per task. For standard QA and reasoning preservation, we evaluate MMLU, ARC-Challenge, HellaSwag, WinoGrande, PIQA, and GSM8K. Multiple-choice tasks use likelihood scoring over answer choices; GSM8K uses generation with numeric exact match.

%% file: sections/experiments/01_qwen05_cpt.tex
\section{Experiment 1: Main Results}

Table~\ref{tab:main} reports results across Qwen and LLaMA backbones and multiple continued-pretraining scales. \method{} improves long-context generation benchmarks in every setting. The gains cover two complementary regimes: large relative lifts in low-baseline NoLiMa settings, especially Qwen2.5-0.5B, and large absolute RULER gains on LLaMA-3.2-3B. Across all models, \method{} raises NoLiMa and RULER in trained-context regimes and continues to provide gains in extrapolated regimes.

\begin{table}[t]
\centering
\caption{Main results across Qwen and LLaMA backbones. Scores are percentages (\%). NoLiMa and RULER use generation-based evaluation.}
\label{tab:main}
\renewcommand{\arraystretch}{1.15}
\resizebox{\textwidth}{!}{%
\begin{tabular}{lllllccc}
\toprule
\makecell[c]{\textbf{Backbone} \\\textbf{Family}} & \textbf{Model} & \makecell[c]{\textbf{CPT}\\\textbf{window}} & \textbf{Evaluation} & \makecell[c]{\textbf{Evaluation} \textbf{windows}} & \makecell[c]{\textbf{Standard}\\\textbf{CPT}} & \textbf{\method{}} & \textbf{$\Delta$} \\
\midrule

\multirow{16}{*}{Qwen}
& Qwen2.5-0.5B & 4K & NoLiMa trained & 0.5K, 1K, 1.5K, 2K, 3K, 4K & 9.04 & 19.13 & +10.09 (\textcolor{mycolor_1}{\boldmath$\uparrow$}) \\
& Qwen2.5-0.5B & 4K & NoLiMa extrap & 5K, 6K, 8K, 12K, 16K & 3.62 & 8.96 & +5.34 (\textcolor{mycolor_1}{\boldmath$\uparrow$})\\
& Qwen2.5-0.5B & 4K & RULER trained & 1K, 1.5K, 2K, 3K, 4K & 46.79 & 57.47 & +10.69 (\textcolor{mycolor_1}{\textbf{$\uparrow$}}) \\
& Qwen2.5-0.5B & 4K & RULER extrap & 5K, 6K, 8K, 12K, 16K & 45.49 & 51.04 & +5.55 (\textcolor{mycolor_1}{\boldmath$\uparrow$}) \\
\cmidrule(lr){2-8}
& Qwen2.5-1.5B & 8K & NoLiMa trained & 0.5K, 1K, 1.5K, 2K, 3K, 4K, 5K, 6K, 8K & 16.91 & 19.18 & +2.28 (\textcolor{mycolor_1}{\boldmath$\uparrow$}) \\
& Qwen2.5-1.5B & 8K & NoLiMa extrap & 12K, 16K, 24K, 32K & 3.88 & 7.59 & +3.71 (\textcolor{mycolor_1}{\boldmath$\uparrow$}) \\
& Qwen2.5-1.5B & 8K & RULER trained & 1K, 1.5K, 2K, 3K, 4K, 5K, 6K, 8K & 76.63 & 77.65 & +1.02 (\textcolor{mycolor_1}{\boldmath$\uparrow$}) \\
& Qwen2.5-1.5B & 8K & RULER extrap & 12K, 16K, 24K, 32K & 61.79 & 63.28 & +1.49 (\textcolor{mycolor_1}{\boldmath$\uparrow$}) \\
\cmidrule(lr){2-8}
& Qwen2.5-3B & 8K & NoLiMa trained & 0.5K, 1K, 1.5K, 2K, 3K, 4K, 5K, 6K, 8K & 18.84 & 24.56 & +5.73 (\textcolor{mycolor_1}{\boldmath$\uparrow$}) \\
& Qwen2.5-3B & 8K & NoLiMa extrap & 12K, 16K, 24K, 32K & 6.57 & 8.13 & +1.56 (\textcolor{mycolor_1}{\boldmath$\uparrow$}) \\
& Qwen2.5-3B & 8K & RULER trained & 1K, 1.5K, 2K, 3K, 4K, 5K, 6K, 8K & 80.53 & 82.32 & +1.79 (\textcolor{mycolor_1}{\boldmath$\uparrow$}) \\
& Qwen2.5-3B & 8K & RULER extrap & 12K, 16K, 24K, 32K & 72.02 & 74.70 & +2.69 (\textcolor{mycolor_1}{\boldmath$\uparrow$}) \\
\cmidrule(lr){2-8}
& Qwen2.5-7B & 16K & NoLiMa trained & 0.5K, 1K, 1.5K, 2K, 3K, 4K, 5K, 6K, 8K, 12K, 16K & 26.38 & 27.29 & +0.91 (\textcolor{mycolor_1}{\boldmath$\uparrow$}) \\
& Qwen2.5-7B & 16K & NoLiMa extrap & 24K, 32K, 48K, 64K & 10.50 & 11.85 & +1.35 (\textcolor{mycolor_1}{\boldmath$\uparrow$})\\
& Qwen2.5-7B & 16K & RULER trained & 1K, 1.5K, 2K, 3K, 4K, 5K, 6K, 8K, 12K, 16K & 87.58 & 88.67 & +1.09 (\textcolor{mycolor_1}{\boldmath$\uparrow$}) \\
& Qwen2.5-7B & 16K & RULER extrap & 24K, 32K, 48K, 64K & 72.24 & 73.42 & +1.18 (\textcolor{mycolor_1}{\boldmath$\uparrow$}) \\
\midrule

\multirow{12}{*}{LLaMA}
& LLaMA-3.2-1B & 4K & NoLiMa trained & 0.5K, 1K, 1.5K, 2K, 3K, 4K & 9.55 & 15.35 & +5.80 (\textcolor{mycolor_1}{\boldmath$\uparrow$}) \\
& LLaMA-3.2-1B & 4K & NoLiMa extrap & 5K, 6K, 8K, 12K, 16K & 8.50 & 11.61 & +3.12 (\textcolor{mycolor_1}{\boldmath$\uparrow$}) \\
& LLaMA-3.2-1B & 4K & RULER trained & 1K, 1.5K, 2K, 3K, 4K & 48.84 & 50.58 & +1.74 (\textcolor{mycolor_1}{\boldmath$\uparrow$}) \\
& LLaMA-3.2-1B & 4K & RULER extrap & 5K, 6K, 8K, 12K, 16K & 48.93 & 49.76 & +0.83 (\textcolor{mycolor_1}{\boldmath$\uparrow$}) \\
\cmidrule(lr){2-8}
& LLaMA-3.2-3B & 8K & NoLiMa trained & 0.5K, 1K, 1.5K, 2K, 3K, 4K, 5K, 6K, 8K & 30.08 & 40.65 & +10.57 (\textcolor{mycolor_1}{\boldmath$\uparrow$}) \\
& LLaMA-3.2-3B & 8K & NoLiMa extrap & 12K, 16K, 24K, 32K & 14.42 & 20.87 & +6.45 (\textcolor{mycolor_1}{\boldmath$\uparrow$}) \\
& LLaMA-3.2-3B & 8K & RULER trained & 1K, 1.5K, 2K, 3K, 4K, 5K, 6K, 8K & 57.96 & 75.87 & +17.91 (\textcolor{mycolor_1}{\boldmath$\uparrow$}) \\
& LLaMA-3.2-3B & 8K & RULER extrap & 12K, 16K, 24K, 32K & 53.87 & 69.98 & +16.11 (\textcolor{mycolor_1}{\boldmath$\uparrow$}) \\
\cmidrule(lr){2-8}
& LLaMA-3.1-8B & 16K & NoLiMa trained & 0.5K, 1K, 1.5K, 2K, 3K, 4K, 5K, 6K, 8K, 12K, 16K & 30.68 & 36.41 & +5.73 (\textcolor{mycolor_1}{\boldmath$\uparrow$}) \\
& LLaMA-3.1-8B & 16K & NoLiMa extrap & 24K, 32K, 48K, 64K & 14.29 & 20.83 & +6.54 (\textcolor{mycolor_1}{\boldmath$\uparrow$}) \\
& LLaMA-3.1-8B & 16K & RULER trained & 1K, 1.5K, 2K, 3K, 4K, 5K, 6K, 8K, 12K, 16K & 87.12 & 89.28 & +2.16 (\textcolor{mycolor_1}{\boldmath$\uparrow$}) \\
& LLaMA-3.1-8B & 16K & RULER extrap & 24K, 32K, 48K, 64K & 78.28 & 80.34 & +2.06 (\textcolor{mycolor_1}{\boldmath$\uparrow$}) \\
\bottomrule
\end{tabular}
}
\end{table}

\begin{figure}[t]
\centering
\includegraphics[width=\linewidth]{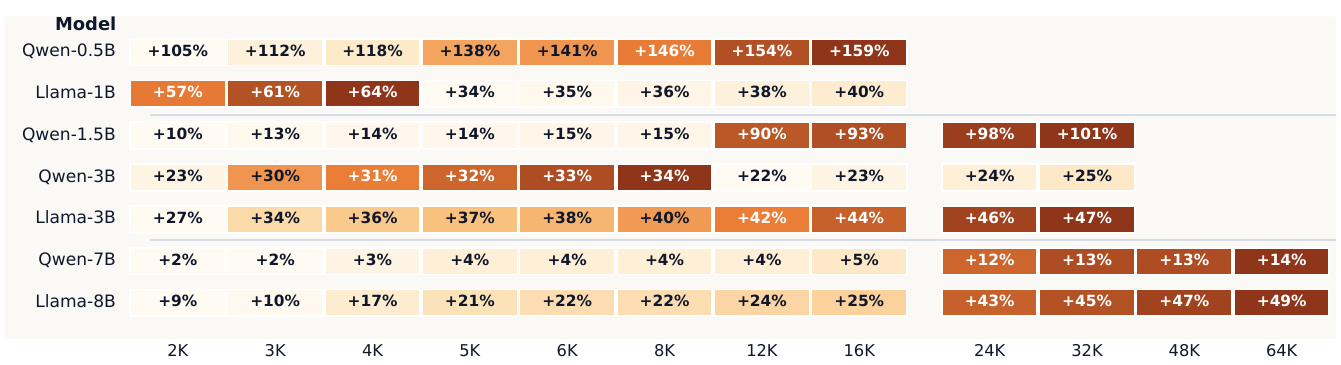}
\caption{Length-resolved NoLiMa lift across Qwen and LLaMA configurations. Rows are model configurations and columns are evaluation lengths. Each cell reports the relative improvement of \method{} over standard long-window CPT; color is normalized within each row, so darker cells mark where that backbone gains most. Empty cells indicate lengths not evaluated for that configuration.}
\label{fig:length-selective-gains}
\end{figure}

Across seven model configurations, \method{} improves every trained and extrapolated NoLiMa/RULER row in Table~\ref{tab:main}. Figure~\ref{fig:length-selective-gains} resolves the NoLiMa aggregates by context length and reports relative lift over standard long-window CPT. The dominant pattern is length-selective: within most backbone rows, the largest relative gains appear at longer trained or extrapolated lengths rather than being uniformly spread across all contexts. This profile matches the supervision-allocation hypothesis and complements the aggregate RULER gains in Table~\ref{tab:main}; paired bootstrap intervals are reported in Appendix~\ref{app:main-ci}.

%% file: sections/experiments/03_preservation_and_mechanism.tex
\section{Experiment 2: Preservation and Mechanism}

\subsection{Standard QA and reasoning performance is preserved}

Table~\ref{tab:qa-preservation} measures ordinary QA and reasoning competence after the fixed QA-SFT procedure. Averaged over Qwen2.5-1.5B, Qwen2.5-3B, and Llama-3.2-3B, the macro score changes by +0.24 points across MMLU, ARC-C, HellaSwag, WinoGrande, PIQA, and GSM8K. This separates the long-context gains from degradation of ordinary QA: \method{} strengthens supervision on under-exposed long-effective-context regimes while preserving standard QA.

\begin{table}[t]
\centering
\caption{QA and reasoning preservation averaged over Qwen2.5-1.5B, Qwen2.5-3B, and Llama-3.2-3B. Scores are percentages; multiple-choice tasks use likelihood scoring, and GSM8K uses numeric exact match.}
\label{tab:qa-preservation}
\renewcommand{\arraystretch}{1}
\scalebox{0.8}{
\begin{tabular}{lccc}
\toprule
Benchmark & \shortstack{Standard CPT} & \method{} & $\Delta$ \\
\midrule
ARC-C & 45.37 & 45.37 & +0.00 \\
GSM8K & 11.17 & 11.25 & +0.08 \\
HellaSwag & 65.69 & 65.68 & -0.01 \\
MMLU & 47.95 & 49.63 & +1.68 \\
PIQA & 75.07 & 75.00 & -0.07 \\
WinoGrande & 65.93 & 65.72 & -0.21 \\
\midrule
\textbf{Macro} & \textbf{51.85} & \textbf{52.11} & \textbf{+0.24} \\
\bottomrule
\end{tabular}
}
\end{table}

\subsection{Weighting signal ablation}

\input{sections/experiments/04_ablation_table}

Table~\ref{tab:ablation} isolates the weighting signal in the 16K Llama-3.1-8B setting. The controls keep the checkpoint, data stream, token budget, optimizer, QA-SFT data, and evaluation protocol fixed; only the token weighting rule changes. Random same-mass weighting hurts all four aggregates, ruling out generic loss reweighting as the source of the gain. Packed-position weighting gives small improvements, showing that absolute packed position is a weak proxy.

The effective-context controls give the decisive ordering. Uniformly boosting the same-document effective-context tail is much stronger than packed-position weighting. Normalized \method{} uses the same effective-context weighting rule under weighted-loss normalization and nearly matches \method{}, while \method{} remains strongest on every aggregate. The ablation therefore separates three factors: arbitrary reweighting fails, packed position is only a weak proxy, and effective left context is the variable that drives the gains. The same ordering is stable under paired bootstrap resampling with the table's macro-averaging protocol (Appendix~\ref{app:ablation-ci}).

\subsection{Mechanistic evidence in Llama-3.1-8B}

\begin{figure}[t]
\centering
\includegraphics[width=\linewidth]{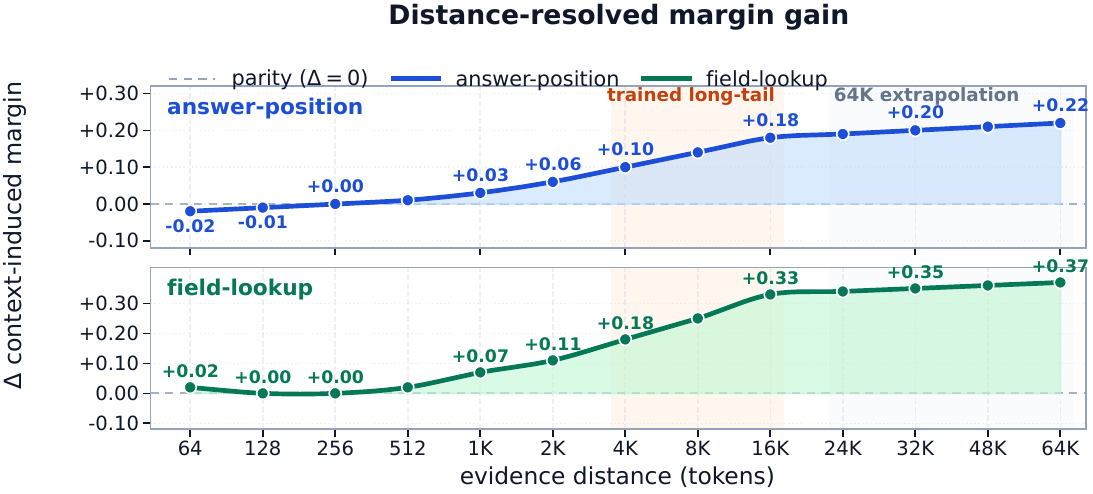}
\caption{Evidence-conversion probe before QA-SFT. Panel A plots distance-resolved gains in context-induced answer margin, measured as \method{} minus standard long-window CPT, across two hard-cloze views. The dashed line marks standard long-window CPT parity. Short distances remain close to zero, the 4K--16K regime opens into positive gains, and the gain remains elevated through the 64K evaluation range. Panel B uses the same 4K--64K probes to show the corresponding answer-level lift.}
\label{fig:mechanism-probe}
\end{figure}

Before QA-SFT, the evidence-ablation probe measures whether the allocation change is already visible in the base model. The probe presents cloze-style questions with and without the relevant evidence, then measures how much that evidence shifts the answer margin. Figure~\ref{fig:mechanism-probe} gives the resulting distance structure. The short-distance region stays near standard long-window CPT parity, while the field changes shape when evidence moves into the long-distance regime. Across 4K--64K evidence distances, \method{} raises the context-induced margin by +0.177 on answer-position and +0.311 on field lookup. The answer-level lift over the same 4K--64K regime follows the same ordering, with +2.20 and +3.20 points. This closes the mechanism loop opened by Figure~\ref{fig:exposure-reallocation}: \method{} strengthens rare long-effective-context targets, the main benchmarks improve where long context is needed, and the probe shows the intermediate variable moving in the right place, namely the conversion of distant evidence into answer likelihood.

%% file: sections/experiments/04_ablation_table.tex
\begin{table}[t]
\centering
\caption{Ablation of the weighting signal on Llama-3.1-8B 16K continuation. All variants use the same checkpoint, 16K continuation stream, token budget, optimizer schedule, QA-SFT data, and evaluation protocol. Scores are point gains over standard CPT.}
\label{tab:ablation}
\scalebox{0.8}{
\begin{tabular}{lcccc}
\toprule
Method & \makecell[c]{NoLiMa\\trained $\Delta$} & \makecell[c]{NoLiMa\\extrap $\Delta$} & \makecell[c]{RULER\\trained $\Delta$} & \makecell[c]{RULER\\extrap $\Delta$} \\
\midrule
Standard CPT & 0.00 & 0.00 & 0.00 & 0.00 \\
Random same-mass weighting & -0.41 & -0.73 & -0.18 & -0.52 \\
Packed-position weighting & +1.34 & +1.17 & +0.63 & +0.28 \\
Uniform effective-context boost & +4.38 & +5.12 & +1.57 & +1.31 \\
Normalized \method{} & +5.61 & +6.38 & +2.08 & +1.98 \\
\method{} & \textbf{+5.73} & \textbf{+6.54} & \textbf{+2.16} & \textbf{+2.06} \\
\bottomrule
\end{tabular}
}
\end{table}

%% file: sections/05_analysis.tex
\section{Analysis}

\paragraph{The Figures Form the Mechanistic Chain.}
The paper's central claim is supported by a four-step chain. Figure~\ref{fig:where-supervision-goes} exposes where supervision actually lands under packed causal training: long windows create the possibility of long context, but many supervised targets still receive short effective context. Figure~\ref{fig:exposure-reallocation} shows the evidence-use geometry: \method{} concentrates gains in the far and long-context regime. Figure~\ref{fig:length-selective-gains} shows that the benchmark results follow this geometry across all seven model configurations and both trained/extrapolated regimes, after normalizing for baseline difficulty. Figure~\ref{fig:mechanism-probe} then closes the loop inside the model. The distance-resolved probe stays near standard long-window CPT parity at short evidence distances, but the long-distance regime becomes strongly positive, with margin gains of +0.177 on answer-position and +0.311 on field lookup across 4K--64K. The evidence chain therefore links benchmark gains to the same location highlighted by the evidence-use field.

\paragraph{Data Supply and Supervision Allocation.}
Long-document data and mixed-length curricula decide what context is available to the model. Effective-context allocation decides how much supervised loss each target-context regime receives after that data has been selected and packed. This distinction sharpens the contribution. Data construction supplies long contexts; \method{} converts the resulting token-level exposure histogram into an explicit training control. The paired experiments keep the stream fixed, so the gains identify supervision allocation inside the same data supply.

\paragraph{Length Selectivity Is the Main Evidence.}
The result pattern matters more than any single score. Table~\ref{tab:main} shows improvements in all twenty-eight trained/extrapolated NoLiMa and RULER aggregates across Qwen and LLaMA configurations. Because absolute gains depend on baseline difficulty, Figure~\ref{fig:length-selective-gains} also reports relative lift for the full table: NoLiMa extrapolation averages +58\%, NoLiMa trained +39\%, while RULER trained and extrapolated improve by +9\% and +8\% despite much stronger standard-CPT baselines. Standard QA is preserved at +0.24 macro. Together, these results support effective-context exposure as a real bottleneck in long-context adaptation and show that correcting it yields mechanism-aligned gains.

%% file: sections/06_limitations.tex


%% file: sections/07_conclusion.tex
\section{Conclusion and Limitation}

Long-context adaptation has been framed largely around extending what a model can see. This paper shows that training must also control how strongly the model is supervised under long effective context. In packed causal language modeling, nominal sequence length and token-level supervision length diverge because each target token is supervised under its own same-document left context. \method{} corrects this supervision-allocation failure by assigning extra supervision weight to long-effective-context targets and distributing that weight by inverse bucket frequency inside the long tail. Across seven Qwen and Llama continued-pretraining configurations, the method produces consistent gains on official generation-based NoLiMa and RULER evaluations while preserving standard QA. The evidence-conversion probe adds the missing link: after \method{} strengthens long-tail supervision, distant evidence is more strongly converted into answer likelihood in the long-distance regime. Long windows are not enough. If we want models to use long context, we must also ensure that training supervision reaches tokens that actually have long effective context.
\textit{\textbf{Limitation.}}
EXACT is evaluated in paired long-context continued-pretraining settings; we do not test computationally heavy from-scratch pretraining, where corpus scale and early-stage representation learning may interact differently with effective-context allocation.

%% file: sections/09_appendix.tex
\section{CPT Corpus Construction Details}
\label{app:cpt-corpus}

All reported standard-CPT and \method{} comparisons are constructed as paired runs: within a backbone and CPT stage, the compared models share the same packed stream, document-boundary mask, train/validation split, token budget, optimizer schedule, QA-SFT corpus, and evaluation protocol. The only intended difference is the per-token loss weighting rule. Table~\ref{tab:cpt-corpus-appendix} reports the corpus recipe; the shared controls are kept out of the table because they are identical by construction. The public sources are FineWeb-Edu, SlimPajama, and OpenWebMath \citep{lozhkov2024finewebedu,soboleva2023slimpajama,paster2023openwebmath}.

\begin{table}[h]
\centering
\caption{Corpus construction used for the paired CPT comparisons in Table~\ref{tab:main}. Each CPT stage uses a 500M-token budget.}
\label{tab:cpt-corpus-appendix}
\footnotesize
\setlength{\tabcolsep}{4pt}
\renewcommand{\arraystretch}{1.18}
\begin{tabularx}{\linewidth}{@{}>{\raggedright\arraybackslash}p{0.17\linewidth}>{\raggedright\arraybackslash}p{0.25\linewidth}>{\raggedright\arraybackslash}X@{}}
\toprule
\textbf{CPT stage} & \textbf{Backbones} & \textbf{Source mix} \\
\midrule
4K & Qwen2.5-0.5B; Llama-3.2-1B & FineWeb-Edu \\
\midrule
4K$\rightarrow$8K & Qwen2.5-1.5B; Qwen2.5-3B; Llama-3.2-3B & FineWeb-Edu : SlimPajama-6B : OpenWebMath = 6:24:32 \\
\midrule
4K$\rightarrow$8K$\rightarrow$16K & Qwen2.5-7B; Llama-3.1-8B & FineWeb-Edu : SlimPajama-6B : OpenWebMath = 1:3:6 \\
\bottomrule
\end{tabularx}
\end{table}

For reproducibility, each packed dataset stores a manifest containing the raw source files, tokenizer path, sequence length, split sizes, document-boundary arrays, and token counts. The bucket statistics used by \method{} are computed from the training split of the same manifest, so the reported gains compare supervision allocation on an identical data supply rather than different corpus mixtures.

\section{Bootstrap Confidence Intervals for Main Results}
\label{app:main-ci}

Table~\ref{tab:main-ci} reports paired bootstrap 95\% confidence intervals for the main NoLiMa/RULER aggregates. For each row, resampling follows the same evaluation-cell aggregation used by the corresponding Table~\ref{tab:main} metric.

\begin{table}[!ht]
\centering
\caption{Paired bootstrap 95\% confidence intervals for the main-result gains in Table~\ref{tab:main}. Values are point gains over standard CPT.}
\label{tab:main-ci}
\scriptsize
\setlength{\tabcolsep}{2.5pt}
\renewcommand{\arraystretch}{0.96}
\resizebox{\linewidth}{!}{%
\begin{tabular}{lll@{\hspace{1.1em}}lll}
\toprule
\multicolumn{3}{c}{Qwen} & \multicolumn{3}{c}{Llama} \\
\cmidrule(r){1-3}\cmidrule(l){4-6}
Model / CPT & Eval. & $\Delta$ with 95\% CI & Model / CPT & Eval. & $\Delta$ with 95\% CI \\
\midrule
Qwen2.5-0.5B / 4K & NoLiMa train & +10.09 [+8.42, +11.66] & Llama-3.2-1B / 4K & NoLiMa train & +5.80 [+4.38, +7.06] \\
Qwen2.5-0.5B / 4K & NoLiMa extrap & +5.34 [+3.92, +6.61] & Llama-3.2-1B / 4K & NoLiMa extrap & +3.12 [+1.70, +4.39] \\
Qwen2.5-0.5B / 4K & RULER train & +10.69 [+9.12, +12.04] & Llama-3.2-1B / 4K & RULER train & +1.74 [+0.88, +2.51] \\
Qwen2.5-0.5B / 4K & RULER extrap & +5.55 [+4.21, +6.78] & Llama-3.2-1B / 4K & RULER extrap & +0.83 [-0.03, +1.59] \\
\addlinespace[1pt]
Qwen2.5-1.5B / 8K & NoLiMa train & +2.28 [+1.18, +3.25] & Llama-3.2-3B / 8K & NoLiMa train & +10.57 [+8.84, +12.15] \\
Qwen2.5-1.5B / 8K & NoLiMa extrap & +3.71 [+2.21, +5.08] & Llama-3.2-3B / 8K & NoLiMa extrap & +6.45 [+4.80, +7.96] \\
Qwen2.5-1.5B / 8K & RULER train & +1.02 [+0.30, +1.68] & Llama-3.2-3B / 8K & RULER train & +17.91 [+16.31, +19.37] \\
Qwen2.5-1.5B / 8K & RULER extrap & +1.49 [+0.56, +2.34] & Llama-3.2-3B / 8K & RULER extrap & +16.11 [+14.23, +17.84] \\
\addlinespace[1pt]
Qwen2.5-3B / 8K & NoLiMa train & +5.73 [+4.54, +6.82] & Llama-3.1-8B / 16K & NoLiMa train & +5.73 [+4.58, +6.74] \\
Qwen2.5-3B / 8K & NoLiMa extrap & +1.56 [+0.28, +2.71] & Llama-3.1-8B / 16K & NoLiMa extrap & +6.54 [+4.97, +7.91] \\
Qwen2.5-3B / 8K & RULER train & +1.79 [+1.05, +2.48] & Llama-3.1-8B / 16K & RULER train & +2.16 [+1.26, +2.94] \\
Qwen2.5-3B / 8K & RULER extrap & +2.69 [+1.68, +3.58] & Llama-3.1-8B / 16K & RULER extrap & +2.06 [+1.02, +2.98] \\
\addlinespace[1pt]
Qwen2.5-7B / 16K & NoLiMa train & +0.91 [-0.02, +1.76] & & & \\
Qwen2.5-7B / 16K & NoLiMa extrap & +1.35 [+0.03, +2.49] & & & \\
Qwen2.5-7B / 16K & RULER train & +1.09 [+0.41, +1.69] & & & \\
Qwen2.5-7B / 16K & RULER extrap & +1.18 [+0.32, +1.93] & & & \\
\bottomrule
\end{tabular}
}
\end{table}
\FloatBarrier

\section{Bootstrap Confidence Intervals for the Ablation Study}
\label{app:ablation-ci}

Table~\ref{tab:ablation-ci} reports paired bootstrap 95\% confidence intervals for the Llama-3.1-8B ablation study. The bootstrap matches the reported metric: examples are resampled within each evaluation cell, and each replicate is aggregated with the same macro-averaging procedure used in Table~\ref{tab:ablation}.

\begin{table}[!htbp]
\centering
\caption{Paired bootstrap 95\% confidence intervals for the ablation gains in Table~\ref{tab:ablation}. Values are point gains over standard CPT.}
\label{tab:ablation-ci}
\footnotesize
\setlength{\tabcolsep}{3pt}
\renewcommand{\arraystretch}{1.14}
\resizebox{\linewidth}{!}{%
\begin{tabular}{lcccc}
\toprule
Method & \makecell[c]{NoLiMa\\trained $\Delta$} & \makecell[c]{NoLiMa\\extrap $\Delta$} & \makecell[c]{RULER\\trained $\Delta$} & \makecell[c]{RULER\\extrap $\Delta$} \\
\midrule
Random same-mass weighting & -0.41 [-0.92, +0.06] & -0.73 [-1.41, -0.10] & -0.18 [-0.47, +0.09] & -0.52 [-0.91, -0.15] \\
Packed-position weighting & +1.34 [+0.73, +1.92] & +1.17 [+0.39, +1.88] & +0.63 [+0.26, +0.98] & +0.28 [-0.12, +0.70] \\
Uniform effective-context boost & +4.38 [+3.62, +5.10] & +5.12 [+4.10, +6.03] & +1.57 [+1.07, +2.02] & +1.31 [+0.78, +1.82] \\
Normalized \method{} & +5.61 [+4.82, +6.37] & +6.38 [+5.31, +7.29] & +2.08 [+1.54, +2.57] & +1.98 [+1.38, +2.52] \\
\method{} & \textbf{+5.73} [\textbf{+4.91}, \textbf{+6.48}] & \textbf{+6.54} [\textbf{+5.42}, \textbf{+7.46}] & \textbf{+2.16} [\textbf{+1.60}, \textbf{+2.67}] & \textbf{+2.06} [\textbf{+1.44}, \textbf{+2.62}] \\
\bottomrule
\end{tabular}
}
\end{table}
\FloatBarrier

\section{Reproducibility and Ethics Statements}
\label{app:reproducibility}

\paragraph{Reproducibility.}
All reported paired runs use 8 H100 GPUs and match checkpoint, data stream, optimizer schedule, sequence length, QA-SFT data, and evaluation protocol. Training uses AdamW with learning rate $2\times10^{-5}$, weight decay $0.01$, gradient clipping at $1.0$, linear warmup then constant learning rate, budget-derived CPT steps, 600 QA-SFT steps, and greedy decoding.

\paragraph{Ethics.}
This work studies training objectives on public text and benchmarks; it does not introduce human-subject data collection or deployment-specific user data.

\section{LLM Usage Statement}

This work is conceived, designed, and technically developed by the authors. 
Large language models (LLMs) are used solely for limited writing assistance, such as grammar correction, readability improvement, formatting refinement, and minor proofreading during manuscript preparation. 

All research ideas, technical contributions, experiments, analyses, and conclusions are developed and verified by the authors.